\documentclass{article}
\usepackage{natbib}
\PassOptionsToPackage{numbers, compress}{natbib}
\usepackage{graphicx}
\usepackage{subcaption}
\usepackage[utf8]{inputenc} % allow utf-8 input
\usepackage[T1]{fontenc}    % use 8-bit T1 fonts
\usepackage{url}            % simple URL typesetting
\usepackage{booktabs}       % professional-quality tables
\usepackage{amsfonts}       % blackboard math symbols
\usepackage{nicefrac}       % compact symbols for 1/2, etc.
\usepackage{microtype}      % microtypography
\usepackage{xcolor}         % colors
\definecolor{darkblue}{rgb}{0, 0, 0.5}
\usepackage[colorlinks=true, linkcolor=darkblue, citecolor=darkblue, urlcolor=darkblue]{hyperref}
\usepackage{amsmath}
\usepackage{amssymb}
\usepackage{mathtools}
\usepackage{amsthm}
\usepackage{bbm}
\usepackage{tcolorbox}
\tcbuselibrary{breakable}
\usepackage[margin=1in]{geometry}

\newcommand{\myquote}[1]{``#1''}

\title{LLM Sparsity Prior for Robust Feature Selection}

\author{%
  \textbf{Caleb Skinner}$^1$, \textbf{Yihan Guo}$^2$, \textbf{Meng Li}$^1$ \\
  $^1$Department of Statistics, Rice University, Houston, TX 77005 \\
  $^2$Department of Computer Science, Rice University, Houston, TX 77005
}

\begin{document}

\maketitle

\begin{abstract}
Large language models (LLMs) offer a scalable mechanism to elicit domain-informed prior information for high-dimensional variable selection. However, existing methods such as \textit{LLM-Lasso} are sensitive to weight quality, with performance degrading substantially when LLM-generated weights are inaccurate. To address this challenge, we first introduce a framework for quantifying the quality of LLM-generated weights, enabling rigorous evaluation of LLM-informed methods across varying weight regimes. We then propose the \textbf{LLM Sparsity Prior (LSP)}, which integrates LLM-generated weights into the prior inclusion probabilities of Spike-and-Slab and Spike-and-Slab Lasso models via two interpretable hyperparameters governing global sparsity and weight concentration. Hierarchical hyperpriors on these parameters allow the model to dynamically discount uninformative or misleading weights, improving robustness without sacrificing gains when weights are accurate. Finally, we develop principled prompt engineering strategies and validate the method on a private medical dataset studying Acute Kidney Injury. LSP improves prediction accuracy and identifies clinically relevant features missed by the baselines, with robustness to prompt variation and particular effectiveness in low-data regimes.
\end{abstract}

\section{Introduction}\label{section-intro}

The integration of large language models (LLMs) into statistical pipelines has emerged as a significant area of research in machine learning \citep{zhang2025, ratkovic2025}. This trend is driven by the success of pre-trained LLMs across a wide variety of domains and tasks \citep{brown2020, radford2019, manikandan2023}. These models excel at encoding domain knowledge and applying it to complex reasoning tasks \citep{wei2023, lewkowycz2022, suzgun2023}. They are increasingly leveraged to conduct traditional machine learning tasks, including prediction \citep{petroni2019, theodoris2023, cui2024}, regression \citep{dinh2022}, and statistical inference \citep{ratkovic2025}.

Feature selection is a task particularly well-suited for LLM-informed modeling, especially in the context of low-data-regimes where signal is sparse and standard machine learning models may struggle \citep{choi2022, zhang2025}. For example, \citet{choi2022} extract a prior distribution from natural language descriptions to enhance feature selection in low-data regimes, \citet{jeong2025} demonstrate impressive accuracy using only the feature set and a description of the prediction task, and \citet{yang2024} combine classical statistical methods with LLM queries in an iterative approach.

The LLM-Lasso \citep{zhang2025} is the first framework to integrate LLM queries and statistical methodologies into a unified feature selection procedure. In this approach, LLM queries generate feature importance weights that encode contextual information derived from feature metadata. These weights are incorporated into the Lasso penalty term, with the degree of reliance on the weights controlled via cross-validation. However, the variance inherent in cross validation-based hyperparameter selection can lead to an over-dependence on the LLM-generated feature weights. These weights directly modify the objective function, and we demonstrate that inaccurate or \myquote{poor} weights can result in substantial performance degradation.

This paper makes three contributions to the growing field of LLM-assisted machine learning. First, we adapt established feature importance evaluation metrics to quantify the quality of the LLM-generated weights, providing a principled diagnostic tool for analyzing the performance of LLM-informed methods across varying weight quality regimes.

Second, we introduce the LLM Sparsity Prior (LSP), a novel framework integrating feature importance weights into the prior inclusion probabilities of Spike-and-Slab and Spike-and-Slab Lasso models. The method utilizes two interpretable hyperparameters: a \textit{sparsity} parameter controlling the baseline inclusion probability and a \textit{concentration} parameter regulating weight influence. We propose hierarchical hyperpriors for both parameters that ensure LSP's robustness to poor weights, while preserving efficient posterior estimation. Simulation studies confirm that LSP remains robust under uninformative or misleading weights, while significantly outperforming baselines when weights are accurate.

Third, we develop principled prompt engineering strategies for querying weights and evaluate the generated weights on a private medical dataset studying Acute Kidney Injury. LSP demonstrates robustness to deviations from the recommended prompt engineering strategy and to the natural stochasticity of LLMs. We find that the proposed method improves the performance of both the Spike-and-Slab and Spike-and-Slab Lasso, particularly in low-data regimes.

The remainder of this paper is organized as follows. Section \ref{section-weight-quality} details the weight quality framework. Section \ref{section-llm-prior-ss} introduces the LLM Sparsity Prior, hierarchical priors, and posterior estimation strategies. Section \ref{section-sims} presents the simulation results, followed by an application to Acute Kidney Injury (AKI) in Section \ref{section-aki}. We conclude the paper in Section \ref{section-conclusion}.

\section{Measuring Weight Quality}\label{section-weight-quality}

Given the proliferation of available LLMs and vast design space of prompting strategies, the variety of generated feature weights is large. LLMs are susceptible to hallucinations and sensitive to prompt variations, thus the derived weights may not accurately capture the ground truth. To assess the performance and robustness of LLM-informed methods to various weights, it becomes necessary to quantify the quality of these weights prior to their integration into statistical pipelines. These quality measurements facilitate rigorous comparison between LLM generation mechanisms and are essential for establishing the statistical properties of LLM-informed methods. We require a weight-quality metric that monotonically increases as weight quality improves and has the precision to permit straightforward comparisons. 

Let $w \in \mathbbm{R}^p_{>0}$ denote a generated weight vector encoding information about $p$ features, where $\text{max}(w) > \text{min}(w)$. In practice, the elements $w_j$ often take values from a discrete set of positive integers, where a larger magnitude suggests a higher probability of inclusion for feature $j$. We define $\gamma^* \in \{0, 1\}^p$ as the ground truth feature inclusion vector such that $\gamma^* = 1$ denotes an active feature in the underlying model and $\gamma^* = 0$ denotes an inactive one. Drawing on established feature importance evaluation practices \citep{heuss2025, catav2021}, we adapt two standard metrics to this setting: $\phi_{\ell_1}, \phi_{\text{pairwise}}: \{0, 1\}^p \times \mathbbm{R}^p_{>0} \to [0, 1]$. For each technique, as $\phi \to 1$, the weight quality approaches perfect alignment, whereas $\phi \to 0$ implies the weights directly oppose the ground truth.

The $\ell_1$ weight agreement is defined
\begin{equation}\label{l1-weight-agreement}
    \phi_{\ell_1}(\gamma^*, w) = 1 - \frac{1}{p}\bigg\|\frac{w - \text{min}(w)}{\text{max}(w) - \text{min}(w)} - \gamma^* \bigg\|_1.
\end{equation}
This metric computes the $\ell_1$ distance between the ground truth and the min-max scaled weight vector. Notably, a value of $\phi_{\ell_1}(\gamma^*, w) \approx 0.5$ indicates that the weights contain no material information regarding $\gamma^*$, equivalent to random guessing.

The pairwise weight agreement measures the frequency with which the relative ordering of elements in $w$ aligns with the strict ordering of $\gamma^{*}$. We define the pairwise indicator matrices for $w$ and $\gamma^{*}$ as
\begin{equation}
    M_{i, j}^{\gamma^{*}} = \mathbbm{1}(\gamma_i^{*} > \gamma_j^{*}) \quad \quad \quad M_{i, j}^{w} = \mathbbm{1}(w_i > w_j).
\end{equation}
The pairwise weight agreement is defined as the normalized sum of disagreements over index pairs:
\begin{equation}\label{pairwise-weight-agreement}
    \phi_{\text{pairwise}}(\gamma^*, w) = 1 - \frac{\sum_{i \neq j} |M_{i, j}^{\gamma^{*}} - M_{i, j}^{w}|}{p(p-1)}.
\end{equation}
Note that $\phi_{\text{pairwise}}$ is related to Kendall's $\tau$ rank correlation, adapted here to the binary ground truth.

In applied settings, evaluating LLM-generated weights prior to deploying LLM-informed methods could guide the selection of specific models or the refinement of prompt strategies. Because the true inclusion vector $\gamma^*$ is unknown, we propose using a standard statistical estimate of the inclusion vector, denoted $\hat{\gamma}$, to construct a natural plug-in estimator: $\hat{\phi} = \phi(\hat{\gamma}, w)$. However, in high-dimensional or low-data regimes, $\hat{\gamma}$ likely serves as a poor estimator, limiting the reliability of inferential conclusions. For this reason, we interpret $\hat{\phi}$ as a measure of the empirical alignment between the weights and the data, rather than a measure of alignment with the true inclusion vector.

\section{LLM Sparsity Prior for Spike-and-Slab}\label{section-llm-prior-ss}

Bayesian prior distributions offer a principled mechanism for integrating LLM-generated weights into statistical models \citep{choi2022}. Ideally, a prior encapsulates expert belief on parameters before the data is observed \citep{deFinetti1979}. However, in high-dimensional regimes, manual elicitation of such priors is infeasible. Consequently, the literature has largely prioritized non-informative or reference priors \citep{jeffreys1946, berger2009}, which express minimal prior knowledge, allowing the data to dominate the posterior distribution.

The emergence of large language models challenges this paradigm, offering a scalable solution for constructing informative priors in high-dimensional contexts. LLMs possess the unique ability to synthesize vast amounts of domain knowledge and contextual metadata into interpretable feature importance weights \citep{zhang2025}. These weights encode \textit{a priori} belief regarding feature relevance, bridging unstructured domain knowledge with formal statistical inference. To maintain the integrity of this Bayesian framework and to prevent data leakage, the LLM must not have access to the training data. This ensures the generated weights represent a genuine \textit{a priori} distribution derived solely from metadata and domain knowledge, rather than an empirical summary of the dataset itself.

In many Bayesian variable selection techniques, the inclusion of the $j = 1, \ldots, p$ features is controlled by latent inclusion variables $\gamma_j$, each governed by a feature-specific inclusion probability $\theta_j$,
\begin{equation}
    \gamma_j|\theta_j \overset{\text{ind}}{\sim} \text{Bernoulli}(\theta_j), \quad j = 1,\ldots, p,
\end{equation}
where $\theta_j$ is the \textit{a priori} probability that the $j^{\text{th}}$ feature is included in the model. In standard applications, analysts typically assume a uniform inclusion probability across all features, setting $\theta_1 = \theta_2 = \ldots = \theta_p$. However, large language models may now provide distinguishing \textit{a priori} information regarding feature relevance.

A naive approach is to instruct a large language model to directly estimate $\{\theta_j\}^p_{j=1}$. However, this requires the LLM to infer both the sparsity of the model and the relative importance of the features. While LLMs are effective at synthesizing information, they can struggle to precisely estimate numerical quantities \citep{yuchi2026}. These concerns are borne out in our data application in Section \ref{section-aki}, where the naive LLM approach is consistently outperformed by non-LLM baselines.

Rather than estimating continuous probabilities, we ask the LLM to assign ordinal importance rankings $w \in \{1, \ldots, K\}^p$. This leverages the LLM's comparative strengths, directing it to perform a simpler categorization task, assessing relative feature relevance, rather than estimating precise numerical quantities. To incorporate the LLM-generated weights while flexibly accounting for the global sparsity, we introduce the \textit{LLM-Sparsity Prior} (LSP). We define each $\theta_j$ in terms of the scaled LLM-generated weight:
\begin{equation}\label{llm-sparsity-prior}
    \theta_j = s \frac{w_j^\eta}{\frac{1}{p}\sum_{k=1}^p w_k^\eta},
\end{equation}
where $s \in (0, 1)$ is baseline sparsity and $\eta \in (0, \eta_{\max})$ controls the degree of contrast between the weights.

The hyperparameters in \eqref{llm-sparsity-prior} offer straightforward interpretation. The parameter $s$ represents the global sparsity level typically employed in standard Spike-and-Slab formulations, while $\eta$ dictates the relative contrast between the weights. We observe that as $\eta \to 0$, the method recovers the traditional setting where $\theta_j = s$ for all $j$. A large value (e.g. $\eta = 10$) sharply amplifies the differences in weight values, while a small value (e.g. $\eta = 0.1$) dampens the LLM's relative preferences, uniformly compressing the transformed weights.

Note that to ensure $\theta_j$ remains a valid probability ($\theta_j < 1)$, the bounds of the concentration parameter $\eta_{max}$ can be selected to satisfy the constraint
\begin{equation}\label{eta-bound}
    w_j^{\eta_{max}} < \frac{1}{sp}\sum_{j=1}^p w_j^{\eta_{max}}.
\end{equation}
If $sp < 1$, then $\eta$ is unbounded. For $sp > 1$, a loose bound is $\eta_{max} \cdot \log\left(\frac{\max(w_j)}{\min(w_j)}\right) = \log\left(\frac{p-1}{sp-1}\right)$, while tighter bounds can be derived computationally in a straightforward manner.

\subsection{Hyperparameter Selection}\label{subsection-hyperparameter}
To account for uncertainty in the global sparsity, it is standard practice to assign a hyperprior to the sparsity parameter $s$. We adopt the natural conjugate Beta distribution,
\begin{equation}
    s \sim \text{Beta}(a_s, b_s),
\end{equation}
where the shape parameters $a_s$ and $b_s$ can be selected to reflect the prior belief on the underlying sparsity.

The concentration parameter $\eta$ may be manually selected to reflect the user's confidence in the LLM, estimated via Empirical Bayes techniques, or assigned a hyperprior and learned directly from the data. We suggest a zero-inflated Discrete Uniform prior on $\eta$ for a fully Bayesian formulation:
\begin{equation}
    \eta \sim \pi_0 \delta_0 + (1- \pi_0)\text{DiscreteUniform}(\mathcal{E}) \label{eta-hyperprior},
\end{equation}
where $\delta_0(\cdot)$ denotes a point mass at zero, $\pi_0 \in [0, 1]$ controls the degree of zero-inflation, and $\mathcal{E}$ is a finite grid of positive candidate values controlling the dispersion of the external weights. $\mathcal{E}$ and $\pi_0$ may be set by the user. We suggest $\pi_0 = 0.5$ and setting $\mathcal{E}$ to be a grid of ten equally spaced values over $(0, \eta_{max}]$.

Under this prior elicitation, $\eta$ is dynamically learned by the model. Specifically, if the LLM-generated weights conflict with the data, the model can fall back to the uninformative baseline by setting $\eta = 0$.

\subsection{Spike-and-Slab Methods}\label{subsection-ss-methods}
We integrate the LLM Sparsity Prior into two Bayesian variable selection techniques: Spike-and-Slab and Spike-and-Slab Lasso. The Spike-and-Slab prior is grounded in a rich literature of theoretical and methodological advancements (see \citet{tadesse-vannucci} for a comprehensive treatment). It is widely regarded as a gold standard for Bayesian variable selection \citep{rockova2018}, and has served as a building block for structured settings, including covariate-dependent and graph-structured models \citep{zeng2025}. Spike-and-Slab Lasso is a more recent advancement and demonstrates state-of-the-art predictive performance \citep{rockova-george2018, bai2020}.

Consider the standard linear regression setting with response $Y \in \mathbbm{R}^n$ modeled by the $n \times p$ design matrix $X$,
\begin{equation}
    Y = \alpha + X\beta + \epsilon,
\end{equation}
where $\epsilon \sim N(0, \sigma^2 I)$ represents Gaussian error, $\alpha$ is the intercept, and $\beta$ is the vector of regression coefficients. The classic discrete Spike-and-Slab prior \citep{mitchell-beauchamp, george-mcculloch} is
\begin{equation}
    \beta_j|\sigma^2, \gamma_j \sim (1-\gamma_j)\delta_0(\beta_j) + \gamma_j N(0, \tau \sigma^2), \quad j = 1, \ldots, p,
\end{equation}
where $\delta_0(\cdot)$ denotes a point mass at zero and $\tau$ controls the slab variance. The Spike-and-Slab Lasso \citep{rockova-george2018} is similar:
\begin{equation}
    \beta_j|\gamma_j \sim (1 - \gamma_j) \frac{\lambda_0}{2} e^{-\lambda_0 |\beta_j|} + \gamma_j \frac{\lambda_1}{2}e^{-\lambda_1 |\beta_j|} +,
\end{equation}
where $\lambda_0$ controls the spike penalty and $\lambda_1$ controls the slab penalty. Typically, $\lambda_1$ is fixed to a small positive value and the Maximum a Posteriori (MAP) estimate is derived using a coordinate descent algorithm that iterates over an increasing grid of $\lambda_0$ values. $\lambda_0$ is selected post-hoc using a model selection criterion such as Bayesian Information Criterion (BIC).

\subsection{Posterior Estimation}\label{subsection-posterior-sampling}
The computational tractability of the LLM Sparsity Prior ensures its compatibility with a wide range of efficient posterior estimation techniques developed for Spike-and-Slab models. For the Spike-and-Slab, we implement an adaptation of the \textit{Add-Delete-Swap} (ADS) algorithm \citep{madigan-york}, a Metropolis-Hastings scheme designed for discrete model space exploration.

Our implementation adapts the standard ADS framework in two key respects. First, the Metropolis-Hastings acceptance ratio is modified to account for the heterogeneous prior inclusion probabilities induced by the LLM-generated weights $w$. Second, we leverage the weights to initialize the inclusion vector $\gamma$, thereby accelerating the convergence to the high-probability region of the posterior. Incorporating random $s$ is straightforward by inserting an additional Metropolis-Hastings update.

If $\eta$ is assigned the hyperprior \eqref{eta-hyperprior}, it may be updated directly within the Gibbs Sampler. Because $\eta$ enters the model solely through $\theta_j$, it is conditionally independent of the data given $\gamma$. Thus, the conditional posterior of $\eta$ is proportional to its prior probability multiplied by the likelihood $p(\gamma | \eta)$.
\begin{equation}
    P(\eta  = k| \gamma) \propto \left\{\begin{aligned}
        & \pi_0p(\gamma | \eta) &&\text{if } \eta = 0 \\
        & \frac{(1 - \pi_0)p(\gamma | \eta)}{n_\eta}&&\text{if } \eta \neq 0, \\
    \end{aligned} \right.
\end{equation}
where $n_\eta = |\mathcal{E}|$. Provided $n_\eta$ is small, computing this exact categorical distribution at each MCMC iteration remains highly computationally tractable.

For the Spike-and-Slab Lasso, the coordinate descent algorithm is run $n_\eta + 1$ times to preserve the increasing penalization structure. For each $\lambda_0$, the value of $\eta$ maximizing the log joint posterior $\log p(\beta, s, \eta | Y)$ is selected. As in the standard Spike-and-Slab Lasso, $s$ may be fixed or treated as random and learned dynamically through the algorithm.

\section{Simulations}\label{section-sims}

We conduct a comprehensive simulation study to investigate the efficacy of integrating LLM-generated weights into statistical feature selection. We integrate the proposed \textit{LLM Sparsity Prior} into Spike-and-Slab and Spike-and-Slab Lasso and compare against four benchmarks: the frequentist \textit{LLM-Lasso} \citep{zhang2025} and the \textit{Spike-and-Slab}, \textit{Spike-and-Slab Lasso}, and \textit{Lasso} baselines. By systematically varying the agreement between the generated weights and the true feature inclusion vector, we enable a controlled evaluation of each method's sensitivity to weight quality.

We give $\eta$ the default zero-inflated discrete uniform prior specified in \ref{subsection-hyperparameter} with $\mathcal{E} = \{1, 2, \ldots, 10\}$ to ensure comparability across all weight quality settings. All Bayesian models follow \citet{rockova2018}, letting $s \sim \text{Beta}(a_s = 1, b_s = p)$. For the Spike-and-Slab methods, we adopt weakly informative inverse-gamma priors on the variance components and set the slab precision $\tau = 1$. We draw $30,000$ posterior samples with $5000$ discarded as burn-in via the ADS algorithm. The Spike-and-Slab Lasso methods set $\lambda_1 = 1$, evaluate $\lambda_0$ on an equally spaced grid on $[1, n]$, and select the final $\lambda_0$ using BIC.

For the LLM-Lasso, we adopt the inverse-importance penalty structure. The estimator minimizes the weighted Lasso objective,
\begin{equation}
    \underset{\beta}{\text{min}}\left\{\frac{1}{2} \sum_{i=1}^n (y_i - \beta_0 - x_i^T \beta)^2 + \lambda\sum_{j=1}^p w_j^{-\eta} |\beta_j| \right\},
\end{equation}
where the hyperparameter $\eta \geq 0$ controls the influence of the LLM-generated weights. Specifically, $\eta = 0$ recovers the standard Lasso, while larger values invoke higher reliance on the weights. Consistent with the original implementation \citep{zhang2025}, we select the optimal hyperparameters using a sequential two-stage cross-validation procedure. First, we optimize the weight reliance parameter $\eta$ over 11 values equally spaced in $[0, 10]$. Second, we tune the regularization parameter $\lambda$ over 100 values equally spaced on $\text{log}\lambda \in [-2.27, 2.34]$.

We generate $n$ samples from $Y = X \beta^* + \mathbf{1}_n\alpha^* + \epsilon$, where $X \sim \text{MVN}(0, \Sigma)$ with $\Sigma_{ii} = 1, \Sigma_{ij} = 0.5$ and $\epsilon \sim N(0, I_n)$. We fix $p = 1000$ with $|\gamma^*| = 20$ active features and all non-zero coefficients in $\beta^*$ and $\alpha^*$ set to 1. We examine two low data regimes: $n = 250$ and $n = 100$.

We generate 25 weight vectors $w \in \{1, 2, 3, 4, 5\}^p$ per setting, spanning qualities from random to perfect: $\phi_{\ell_1}(\gamma^*, w) \in \{0.50, 0.60, 0.70, 0.75, 0.80, 0.81, \ldots, 1.00\}$. This fine-grained grid enables precise characterization of each method's sensitivity to weight quality. The generation algorithm is provided in Appendix \ref{subsection-app-weight-generation}.

\begin{figure}[htbp]
    \centering
    \includegraphics[width=0.9\linewidth]{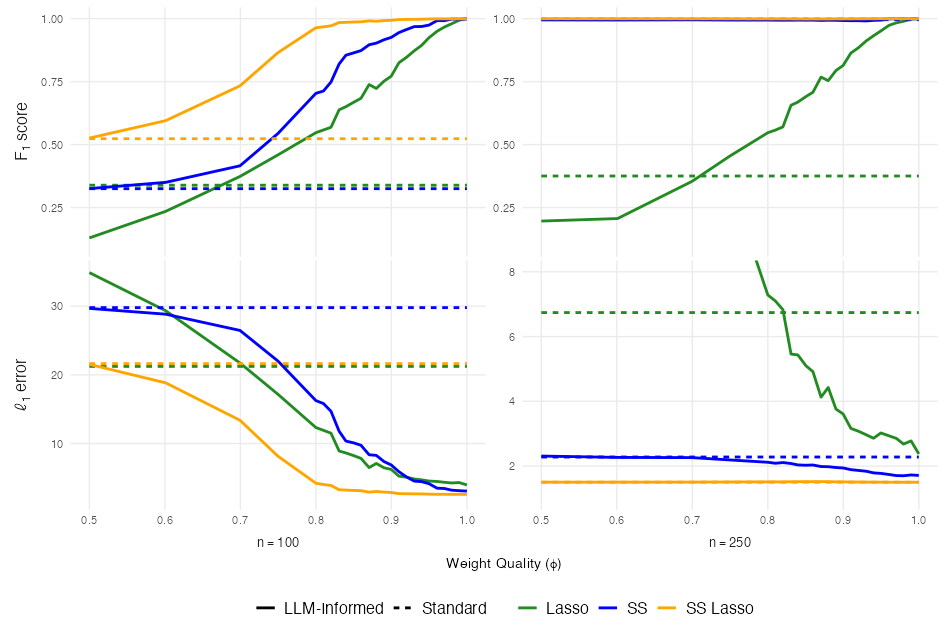}
    \footnotesize
    \raggedright
    \caption{For both $n = 100$ and $n = 250$, LSP is robust, outperforming the respective baseline at all weights and dramatically improving with quality weights. Conversely, LLM-Lasso underperforms the baseline with lower-quality weights.}
    \label{fig:sim-both-n}
\end{figure}

We evaluate the methods on two primary tasks: feature selection accuracy and coefficient structure recovery. For the Spike-and-Slab methods, regression coefficients are estimated via Bayesian Model Averaging \citep{hoeting1999} and the active feature set $\hat{\gamma}$ are estimated via the Median Probability Model \citep{barbieri-berger}. We report the $F_1$ score for feature set recovery and $\ell_1$ error for coefficient estimate precision, averaged across 500 replications in Figure \ref{fig:sim-both-n}.

All three LLM-informed methods improve substantially as the weight quality increases. For example, in the high-dimensional regime, the Spike-and-Slab yields an $\ell_1$ error of 29.78, and the LSP (SS) reduces this error fourfold at $\phi_{\ell_1} = 0.90$. Similarly, the $F_1$ score approaches a perfect score of 1.0 as the weight quality improves, despite the baseline scoring $0.325$. This performance gain is observed across all three LLM-informed methods, and is most pronounced in the low-data regime.

The LLM-informed methods differ considerably in their sensitivity to the weight quality. When integrated with both Spike-and-Slab and Spike-and-Slab Lasso, the LLM Sparsity Prior demonstrates remarkable stability; effectively capitalizing on high-quality information without suffering degradation with poor weights. The performance of the LSP methods is lower bounded by their respective baselines across all weight qualities. Under random weights ($\phi_{\ell_1} = 0.5$), the LSP methods yield $\ell_1$ errors within $0.1$ units of the baseline and $F_1$ scores within $0.005$ units of the baseline.

In contrast, the LLM-Lasso is highly volatile. Under moderate weight regimes ($\phi_{\ell_1} \leq 0.7$), its performance deteriorates severely, yielding estimation error up to 1.64 times more than that of the Lasso. Although the method includes a cross-validation step intended to revert to the standard Lasso when weights are uninformative, our results suggest that \textit{moderate}-quality weights may mislead the cross validation procedure, trapping the algorithm in a suboptimal model.

Analysis of feature inclusion probabilities reveals a sharp contrast in how the methods utilize the feature weights. Focusing on a mediocre weight regime ($n = 100, \phi_{\ell_1}(\gamma^*, w) = 0.70$; Appendix \ref{subsection-app-mip}), we observe that the LLM-Lasso is aggressively directed by the weights. Inactive features assigned high importance ($w_j = 5$) are incorrectly included 35.6\% of the time and active features assigned low importance ($w_j = 1)$ are never included. Conversely, the LSP methods are guided by the weights but not overwhelmed by them.

The selection of $\eta$ and its prior requires careful consideration. For both LSP (SS) and LSP (SSL), we evaluate $\eta$ across an integer grid on $[1, 20]$ and compute the $\ell_1$ structure recovery error, presented in Appendix \ref{subsection-app-eta-sensitivity}. Across two weight qualities $\phi = \{0.8, 0.9\}$, the performance of each method is maximized at a central value of $\eta$, typically $\eta \approx 5$. After this point, the performance slowly decreases. Crucially, the performance achieved at the optimal fixed $\eta$ closely matches that of the zero-inflated Discrete Uniform prior, substantiating the hyperprior.

\section{Application to Acute Kidney Injury}\label{section-aki}

We validate the proposed framework using the BCM Cardiothoracic Surgery EMR Database, an unpublished clinical cohort focused on acute kidney injury (AKI) curated by Baylor College of Medicine (BCM) \citep{ryan2022}. As these data are unpublished, they remain insulated from the LLM's training corpus, ensuring the generated weights constitute genuine \textit{a priori} information. All patients in the database underwent cardiac surgery between 2017 and 2022, but we focus specifically on five subpopulations: patients over 80 years of age, female smokers, black men, persons with liver disease, and immunocompromised individuals. Each of these subsets has between $n = 163$ and $n = 265$ patients. After preprocessing, the feature space consists of between $p = 1063$ and $p = 1079$ predictors, encompassing demographic information, routine laboratory panels, medication records, and hemodynamic metrics collected until 36 hours post-operation. Our objective is to predict the post-operation increase in serum creatinine levels 60 hours  relative to baseline, the primary clinical indicator for AKI.

We generate feature weights using GPT-5.2o \citep{openai2025gpt} via a zero-shot approach with greedy decoding (temperature zero), relying entirely on the model's pre-trained knowledge base to improve reproducibility \citep{radford2019} and preserve the \textit{a priori} nature of the weights. The full prompt, located in Appendix \ref{subsection-app-prompt}, consists of five modules:

\begin{tcolorbox}[
    title=Prompt Engineering Modules,
    fonttitle=\bfseries,
    colback=gray!5,
    colframe=black!60,
    breakable
]
\begin{enumerate}
    \item \textbf{Background.} Clinical context, dataset background, and the temporal forecasting window for the target window are established.
    \item \textbf{Task.} The LLM's primary task is defined, instructing it to evaluate the predictive utility of each feature as a domain expert.
    \item \textbf{Constraints.} The LLM is instructed to identify and penalize redundant EMR artifacts and prioritize physiological drivers.
    \item \textbf{Scoring Rubric.} The model is provided a strict scoring rubric, mapping its clinical assessment to a 1--5 integer scale.
    \item \textbf{Formatting.} Formatting and chain-of-thought directives are specified, guiding the LLM to articulate its rationale.
\end{enumerate}
\end{tcolorbox}
We restrict the model space to Bayesian methods for their interpretability and uncertainty quantification advantages. We compare the performance of the LSP for Spike-and-Slab and LSP for Spike-and-Slab Lasso with the classic Spike-and-Slab, Spike-and-Slab Lasso, and Horseshoe Prior \citep{carvalho2010}. All hyperparameter settings are maintained from Section \ref{section-sims}, with one exception: we draw $60,000$ posterior samples, discarding the first $10,000$ as burn-in. To ensure robust performance estimates, we conduct five-fold cross-validation, repeated ten times, and report the aggregated out-of-sample Mean-Squared Error (MSE) in Table \ref{tab:aki-dataset}. Full computational details are provided in Appendix \ref{section-app-computational}.

\begin{table}[ht]
    \centering
    \footnotesize
    \raggedright
    \caption{Out-of-Sample Mean Squared Error (multiplied by a factor of $10^2$) for five subsets. Standard Errors across repetitions are reported in the final row.}
    \label{tab:aki-dataset}
    \centering
    \begin{tabular}{cc ccccc}
        \toprule
        Method    & LLM-Weights   & Elderly Patients & Female Smokers & Black Men & Liver Disease & Compromised \\
        \midrule
        SS        & Standard      & 4.34          & 8.00          & 6.83          & 3.66          & 17.57 \\
        SS        & LSP           & \textbf{4.11} & \textbf{7.31} & \textbf{6.29} & \textbf{3.63} & \textbf{17.17} \\
        SS        & Naive Weights & 4.60          & 10.61         & 10.21         & 4.93          & 18.52 \\
        \multicolumn{2}{c}{Maximum SE}    & 0.10          &  0.32         &  0.26         & 0.05          &  1.21\\
        \midrule
        SSL  & Standard      & 3.81          & 7.17          & 6.31          & \textbf{3.49} & 14.30 \\
        SSL  & LSP           & \textbf{3.79} & \textbf{7.04} & 6.29          & 3.61          & 13.71 \\
        SSL  & Naive Weights & 3.82          & 7.20          & \textbf{6.15} & 3.50          & \textbf{13.69} \\
        \multicolumn{2}{c}{Maximum SE}    & 0.04          & 0.34          & 0.10          & 0.04          & 0.74\\
        \midrule
        Horseshoe & Standard      & 4.43          & 8.69          & 8.91          & 3.78          & 17.57 \\
        \bottomrule
    \end{tabular}
\end{table}

The application demonstrates the clear advantages of the LLM Sparsity Prior. LSP (SS) yields an improvement over the baseline Spike-and-Slab in all five subsets, reducing the out-of-sample MSE by an average of 5\% across the subsets. Notably, the naive weights (SS) substantially increase MSE in every subset.

The LSP (SSL) improves the SS Lasso in four of five subsets. The comparatively smaller effect relative to the Spike-and-Slab is mechanistically expected: in highly sparse settings, the spike penalty $\lambda_0$ serves as the primary sparsification mechanism, and the inclusion probability $\theta_j$ plays a secondary role. Consequently, the LLM-generated weights have less leverage over the final model, reducing the benefit of the LSP. The naive LLM weights outperform the standard SS Lasso in two of five subsets, suggesting that in highly sparse settings, the regularization imposed by $\lambda_0$ already captures much of the signal the weights would otherwise provide.

Beyond predictive accuracy, the LSP enhances feature selection. We set $\tau = 2$ and apply both Spike-and-Slab methods to the full dataset, reporting the Marginal Inclusion Probability (MIP) for the top ten features in Appendix \ref{subsection-app-aki}. Using the Median Probability Model \citep{barbieri-berger}, LSP (SS) selects three features: the Maximum Creatinine Ratio over hours 25-36, Intraoperative Red Blood Cells transfused (RBC), and Maximum Creatinine Ratio over hours 13-24. While one of these features is also selected by the baseline Spike-and-Slab, LSP (SS) uniquely identifies RBC, a crucial biomarker omitted by the baseline. Blood transfusions are well-documented predictors of AKI in the nephrology literature and are independently associated with AKI \citep{mendez2024}. Thus, incorporating the LLM-feature weights guides the mechanism toward a more clinically sound solution, recovering a clinically meaningful feature that was missed by the standard method.

\begin{figure}[ht]
    \centering
    \includegraphics[width=0.5\linewidth]{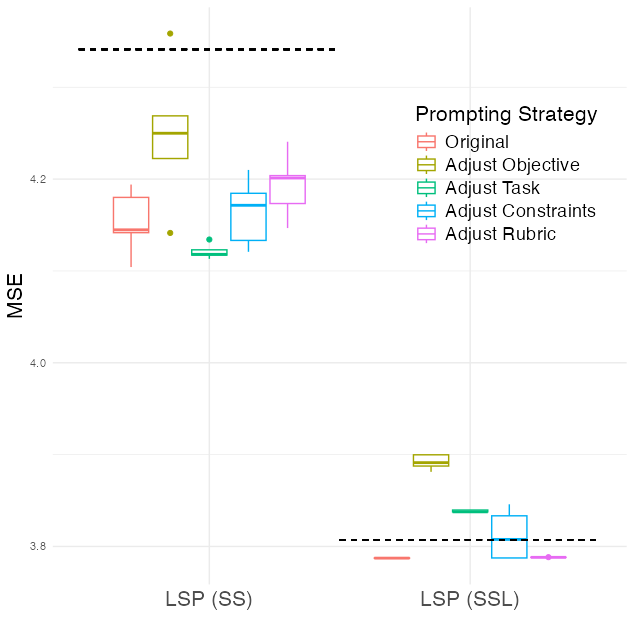}
    \captionof{figure}{MSE over five prompting strategies. Dotted line is the associated baseline.}
    \label{fig:weight-sensitivity}
\end{figure}

Given the stochasticity inherent in LLMs \citep{atil2025} and the enormous prompt design space with few theoretical guarantees to serve as guides, we conduct a sensitivity analysis on the LLM generated weights. We write four new weight-generation prompts, each adjusting one module: the objective and reasoning structure, the task definition, the redundancy constraints, and the scoring rubric (extended to 1--10). For all five prompts, we generate five importance weight vectors and apply them to the elderly patient subset, reporting out-of-sample MSEs in Figure \ref{fig:weight-sensitivity}.

This weight sensitivity exercise reveals several notable findings. LSP (SS) remains robust across the 25 weight vectors, outperforming the baseline in 24 of 25 cases with minimal variation across LLM draws. Results for LSP (SSL) are more mixed: two prompting strategies strictly improve the baseline while two degrade it. Notably, four of the five prompting strategies perform similarly, suggesting reasonable adjustments to the prompt will not dramatically hinder results. The original prompt performs well, and it remains our recommended strategy.

Last, we hypothesize that the impact of the LLM Sparsity Prior will increase as the sample size decreases and the baseline approaches struggle to identify the signal. To study this, we subsample the cross-validation folds of the elderly patient subset, training on data sets of $n \in \{100, 150, 200\}$ and report the out-of-sample MSE in Table \ref{tab:low-data}. Consistent with this hypothesis, the performance gain of the proposed method over the baseline widens as the training sample size decreases for both LSP (SS) and LSP (SSL). This confirms LSP as a particularly effective strategy in low-data regimes.

\begin{table}
    \centering
    \footnotesize
    \captionsetup{width=0.78\linewidth}
    \captionof{table}{Performance Gain (MSE scaled by $10^2$) over baseline at select $n$. Standard Errors over 10 repetitions are in parentheses.}
    \label{tab:low-data}
    \begin{tabular}{l cc}
        \toprule
        Training & LSP (SS) & LSP (SSL)\\
        \midrule
        $n = 100$   & 0.27 (0.13) & 0.16 (0.06)\\
        $n = 150$   & 0.27 (0.06) & 0.06 (0.02)\\
        $n = 200$   & 0.19 (0.07) & 0.06 (0.03)\\
        Full Data   & 0.21 (0.13) & 0.02 (0.02)\\
        \bottomrule
    \end{tabular}
\end{table}

\section{Conclusion}\label{section-conclusion}

In this paper, we propose the LLM Sparsity Prior (LSP), a novel method for integrating LLM-informed feature weights under the Bayesian paradigm. We implement the LSP into the classic Spike-and-Slab and Spike-and-Slab Lasso models, detailing efficient posterior estimation techniques that maintain the flexibility and properties of the baseline methods. While all LLM-informed methods offer substantial gains over the baselines when weights are informative, the LLM Sparsity Prior remains robust to misleading weights. This practical utility is empirically demonstrated in our application to Acute Kidney Injury, where LSP reduces prediction error and recovers critical physiological markers largely missed by the standard Spike-and-Slab. In this application, the proposed method is shown to be robust to several prompting strategies and particularly effective in low-data regimes. Consequently, in real-world applications where weight quality is unknown, LSP offers a reliable and effective alternative.

The extent of this advantage, however, depends on the degree to which prior inclusion probabilities govern variable selection in the underlying model. In Spike-and-Slab Lasso, where the spike penalty $\lambda_0$ serves as the primary sparsification mechanism in highly sparse regimes, the LLM-generated weights have comparatively less leverage over the final model. In the current construction, the coordinate descent over the spike penalty $\lambda_0$ is not informed by the weights. Future work should study this relationship and consider an approach that integrates LLM-weights directly into the coordinate descent of $\lambda_0$.

The zero-inflated discrete uniform prior on $\eta$ is a crucial component of the LSP hierarchy. Placing a large mass on the baseline model ($\eta = 0$) and evaluating over a discrete set maintains computational efficiency and ensures robustness. However, other continuous priors for $\eta$ should be considered. Their implementation would require considerable computational adjustments --- a Metropolis-Hastings step in the Spike-and-Slab and an adjustment of the coordinate descent algorithm in Spike-and-Slab Lasso --- but such extensions merit further investigation.

The performance of LLM-informed methods relies heavily on the quality of LLM-generated weights, yet in practice, the quality is unknown and difficult to measure. While we propose a plug-in estimator for this purpose, future work should develop principled criteria for evaluating LLM-generated weights. Such criteria would also provide an objective basis for comparing and refining prompt engineering strategies. For these LLM-informed methods to become widely used, robust prompting principles must be formalized. Even under a fixed prompt, the stochasticity of LLM outputs introduces variability that motivates techniques for stabilizing the feature importance weights. One natural approach is to generate multiple independent weight vectors and aggregate them via their mean or median.

Finally, the LLM Sparsity Prior framework holds significant potential for broader application in Bayesian regularization. The mechanism extends naturally to other variable selection techniques that employ feature-level inclusion or splitting probabilities, such as Bayesian tree ensembles \citep{chipman2010, linero2018, ye2025posteriorsummaries} and graphical models. Similar ideas may also be useful for LLM-informed sparse scientific discovery, including high-dimensional symbolic regression \citep{ye2025symbolicregression}. Extending the LSP framework to continuous shrinkage priors, such as the Horseshoe Prior \citep{carvalho2010}, requires further methodological development, as these models do not employ explicit inclusion probabilities. Moreover, while this work focuses on regression, extension to classification is natural via logistic or probit regression.

\paragraph{Code and Reproducibility} To facilitate reproducibility, we provide end-to-end implementation at \url{https://github.com/CalebSkinner1/LLMSparsityPrior}.

% Acknowledgements should only appear in the accepted version.
% \section*{Acknowledgments}

%\newpage 

\bibliographystyle{apalike}
\bibliography{references}

\newpage

\appendix

\section{Additional Simulation Notes}\label{section-app-simulation}

\subsection{Simulation Weight Generation}\label{subsection-app-weight-generation}

To systematically vary the quality of feature importance weights across simulations, we implement a probabilistic generation mechanism controlled by the $\ell_1$ weight agreement, $\phi_{\ell_1}$. We generate integer weights $w_j \in \{1, 2, 3, 4, 5\}$ where the \myquote{ideal} LLM would assign the maximum weight $(w_j = 5)$ to active features and the minimum weight $(w_j = 1)$ to inactive features. We introduce noise into this process by modeling the deviation distance $d$ from the ideal weight, where $d \in \{0, 1, 2, 3, 4\}$.

The distribution of these deviations follows geometric decay, so the probability of observing the deviation $d$ is proportional to $r^d$ for some ratio $r \in [0, 1]$. The probability mass function for the deviation is written
\begin{equation*}
    P(D = d) = \frac{r^d}{\sum_{k = 0}^4 r^k} \text{for } d \in \{0, 1, 2, 3, 4\}.
\end{equation*}
The parameter $r$ is calibrated to match the target quality $\phi_{\ell_1}$. We define the target expected deviation $\mu$ as a linear function of $\phi_{\ell_1}$: $\mu = 4(1-\phi_{\ell_1})$. This implies that for perfect weights, the expected deviation is 0. We solve for the unique root $r$ in the polynomial equation derived from the expected value definition,
\begin{equation*}
    (4 - \mu)r^4 + (3 - \mu)r^3 + (2-\mu)r^2 + (1-\mu)r - \mu = 0.
\end{equation*}
After solving for $r$, the probabilities $P(D = d)$ follow. We assign the counts for each weight class for the $p_1 = |\gamma^*|$ active features and $p_0 = p - p_1$ inactive features according to the probabilities $P(D = d)$, rounding to the nearest integers and preserving the total number of features $p$. For example, the count of inactive features with weight $w_j = 1$ is approximately $p_0 \cdot P(D = 0)$, the count of inactive features with weight $w_j = 2$ is approximately $p_0 \cdot P(D = 1)$, etc. This ensures that the generated weight vector $w$ closely approximates the target $\ell_1$ weight agreement, while deviations have geometrically decreasing probabilities.

\subsection{Mean Inclusion Probability}\label{subsection-app-mip}

We report the mean marginal inclusion probabilities under the simulation settings specified in Section \ref{section-sims}. The LSP methods are guided by the weights, but resistant when weights are inaccurate. Conversely, LLM-Lasso is directly controlled by the weights and susceptible to strong errors when weight quality is poor.

\begin{table}[ht]
    \footnotesize
    \caption{Mean feature inclusion probabilities over 500 replications, grouped by weight ($w_j$) and true inclusion status ($\gamma_j$) at $n = 100$ and $\phi_{\ell_1}(\gamma^*, w) = 0.70$. Column headers denote $w_j$ values.}
    \label{tab:sim-inclusion-probability}
    \centering
    \begin{tabular}{l ccccc ccccc}
        \toprule
        & \multicolumn{5}{c}{$\gamma_j = 0$} & \multicolumn{5}{c}{$\gamma_j = 1$}\\
        \cmidrule(lr){2-6} \cmidrule(lr){7-11}
        Method         & 1     & 2     & 3     & 4     & 5     & 1     & 2     & 3     & 4     & 5\\
        \midrule
        LSP (SS)       & 0.008 & 0.011 & 0.015 & 0.022 & 0.034 & 0.202 & 0.265 & 0.319 & 0.366 & 0.412\\
        SS             & 0.015 & 0.015 & 0.015 & 0.015 & 0.015 & 0.272 & 0.272 & 0.272 & 0.272 & 0.272\\
        \midrule
        LSP (SSL)      & 0.003 & 0.004 & 0.005 & 0.006 & 0.009 & 0.567 & 0.636 & 0.690 & 0.721 & 0.750\\
        SSL            & 0.007 & 0.007 & 0.007 & 0.007 & 0.007 & 0.483 & 0.483 & 0.483 & 0.483 & 0.483\\
        \midrule
        LLM-Lasso      & 0.000 & 0.001 & 0.024 & 0.136 & 0.356 & 0.000 & 0.209 & 0.598 & 0.895 & 0.984\\
        Lasso          & 0.070 & 0.070 & 0.070 & 0.070 & 0.070 & 0.897 & 0.897 & 0.897 & 0.897 & 0.897\\
        \bottomrule
    \end{tabular}
\end{table}

\subsection{Concentration Parameter Sensitivity}\label{subsection-app-eta-sensitivity}

We conduct a simulation study on the performance of LSP across fixed values of $\eta$. Under the same settings specified in Section \ref{section-sims}, we generate 500 replications with $n = 100$ samples. We compare the results at each fixed value of $\eta$ with the proposed zero-inflated discrete uniform prior. We find that the proposed prior is near to the maximum performance on the fixed grid.

\begin{figure}[ht]
    \centering
    \includegraphics[width=0.9\linewidth]{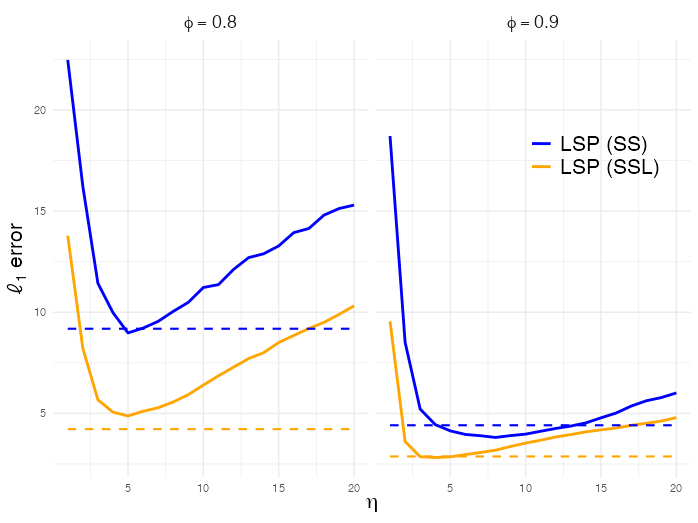}
    \caption{Structure Recovery of LSP Methods across selected $\eta$ values. The horizontal lines denote the zero-inflated prior ($\mathcal{E} = \{1, 2, \ldots, 10\}$), which consistently achieves results comparable to the peak performance of the individual $\eta$ values.}
    \label{fig:eta-sensitivity}
\end{figure}

\newpage
\section{Additional AKI Notes}\label{section-app-aki}

\subsection{Weight Generation Prompt}\label{subsection-app-prompt}

We attach the original prompt below. The four adjusted prompts and the prompt used to elicit continuous inclusion probabilities are located online at \url{https://github.com/CalebSkinner1/LLMSparsityPrior}.

You are a cardiothoracic ICU clinician and a biostatistician.

\textbf{Background:}\\
We have adult patients, 80 years and older, who underwent cardiac surgery. For each patient, we have perioperative and ICU EMR-derived features: demographics, comorbidities, vitals/hemodynamics, ventilator settings, fluid balance, labs, procedures, and medication doses. Features are collected up until 36 hours post-surgery. The modeling goal is sparse linear regression to predict the change in postoperative serum creatinine after 60 hours. Specifically, we want to predict the ratio of postoperative serum creatinine after 60 hours over the patient's baseline (pre-surgery) creatinine levels.

\textbf{Task:}\\
We are building a sparse linear regression model to predict Acute Kidney Injury (AKI) in adult patients, 80 years and older, following cardiac surgery. The exact target variable, \texttt{creatinine\_ratio}, is the patient's postoperative serum creatinine at Hour 60 divided by their baseline (pre-surgery) creatinine.

For each feature, measure its clinical relevance in predicting \texttt{creatinine\_ratio} in post-cardiac surgery patients on an integer scale from 1--5. Base your judgment on typical knowledge about kidney perfusion/hemodynamics, AKI risk factors, nephrotoxic medications, and kidney-related labs.

Rules for Aggregated Vitals and Labs:\\
You will evaluate many features ending in \texttt{\_min}, \texttt{\_max}, \texttt{\_mean}, and \texttt{\_measured} representing a 12-hour post-surgery window. You must differentiate your scores based on these specific aggregations:

\textbf{Collinearity Rule:}\\
We want to remove collinearity in the aggregated features. When making your choices, you must select the aggregation that best captures the true pathology. For example, if a feature is a \texttt{\_mean}, and you know the \texttt{\_max} or \texttt{\_min} captures the true pathology better, strictly cap the feature at a score of 2 to prevent redundancy.

Extremes (\texttt{\_max}, \texttt{\_min}): These capture acute physiological insults. If the extreme state is a known physiological trigger or direct marker for AKI, score it according to the rubric.\\
Averages (\texttt{\_mean}): These smooth out acute results. Unless the 12-hour sustained average is the primary drive of the pathology, reduce your score.\\
Measurement Frequency (\texttt{\_measured}): This represents the number of times a test or measurement was ordered. This is a behavioral proxy for clinical suspicion/acuity, NOT a physiological mechanism. You must score \texttt{\_measured} features as a 1 or 2 unless you have strong belief that this test or measurement indicates a physician's belief that AKI may be imminent.

Rule for Temporal Proximity (Time Epochs):
Many of the features are divided into three consecutive 12-hour epochs (\texttt{0\_12h}, \texttt{13\_24h}, \texttt{25\_36h}) to predict a clinical outcome at hour 60. You must explicitly adjust your scores based on the epoch:

Initial Period (\texttt{0\_12h}): This period captures the immediate trauma of cardiac surgery, anesthesia, and the
cardiopulmonary bypass machine. Derangements are common and transient. Be very conservative with your scores
in this period. Maximum Score: 3.

Trajectory Phase (\texttt{13\_24h}): This period shows whether the patient is stabilizing or deteriorating. Reduce your scores in this period. Maximum Score: 4.

Leading Indicator (\texttt{25\_36h}) : This is the most critical window as it is the closest physiological snapshot to the Hour 60 outcome. Full 1--5 scoring allowed.

\textbf{Scoring rubric:}\\
Use an integer score from 1 to 5:\\
Score 1 (No/Weak Evidence): The feature has no meaningful physiological link to AKI, OR it represents routine, standard-of-care ICU maintenance that provides no specific prognostic value for renal failure.\\
Score 2 (Plausible Indirect Link): There is a theoretical or indirect physiological link, but it is not a primary driver or established predictor of AKI.\\
Score 3 (Established Risk Factor): The feature is a known comorbidity, standard hemodynamic indicator, or medication that routinely influences renal perfusion or AKI risk, but is not definitive on its own.\\
Score 4 (Strong Direct Predictor): Strong clinical evidence links this feature directly to subsequent AKI.\\
Score 5 (Definitive/Direct Biomarker): The feature is a direct, early measurement of renal failure or severe hemodynamic collapse explicitly known to cause renal tubular necrosis.

\textbf{Formatting Directives:}
Return a JSON object with key "scores" containing a list of objects.
Each object must include:
- id (same as input id)
- name (copy exactly)
- importance (integer 1..5)
- reason (1–2 concise sentences)

\subsection{AKI Marginal Inclusion Probabilities}\label{subsection-app-aki}

We report the marginal inclusion probabilities for ten features in Table \ref{tab:aki-mip}. Crucially, LSP (SS) uncovers RBC Transfusions, a key clinical driver of AKI \citep{mendez2024}, while reducing the posterior inclusion probability of noise features like Total Aspirin Dosage (13-24 hours).

\begin{table}[ht]
        \centering
        \footnotesize
        \raggedright
        \caption{Marginal Inclusion Probability for Select AKI Features. We list ten features and their associated LLM-generated weight. LSP (SS) identifies a key driver of AKI (RBC Transfusions) and reduces the posterior inclusion probability of noise variables.}
        \label{tab:aki-mip}
        \centering
        \begin{tabular}{l ccc}
            \toprule
            Feature                                     & $w_j$ & LSP (SS) & SS\\
            \midrule
            Max Creatinine Ratio (25-36 hours)          & 5 & 1.000   & 1.000 \\
            Intraoperative Red Blood Cell Transfusion   & 4 & \textbf{0.602}  & \textbf{0.025} \\
            Max Creatinine Ratio (13-24 hours)          & 4 & 0.526  & 0.018 \\
            Operative Lowest Hemoglobin                 & 4 & 0.266  & 0.001 \\
            Max Phosphorus Levels (25-36 hours)         & 4 & 0.124  & 0.010 \\
            --- \\
            Regular Cardiac Condition (25-36 hours)     & 1 & 0.000  & 0.022 \\
            Min Phosphorus Levels (25-36 hours)         & 1 & 0.000  & 0.016 \\
            Min Creatinine Ratio  (25-36 hours)         & 1 & 0.000  & 0.013 \\
            Mean Phosphorous Levels (25-36 hours)       & 2 & 0.000  & 0.012 \\
            Total Aspirin Dosage (13-24 hours)          & 1 & 0.000  & 0.012 \\
            \bottomrule
        \end{tabular}
\end{table}

\newpage

\section{Computational Details}\label{section-app-computational}
LLM-generated feature weights were obtained via the OpenAI API using GPT-5.2o with greedy decoding; weight generation for each prompt required a single API call of approximately 150,000 tokens. All experiments were implemented in R (version 4.5.2) and executed on a computing cluster using 50 CPU cores with 1.1 GB per core. Each simulation replication required approximately 20 seconds for LSP (SS) and 90 seconds for LSP (SSL), yielding a total runtime of approximately 800 hours across 500 replications, 25 weight quality settings and two sample sizes. Each cross validation fold in the AKI application required approximately 2 minutes, yielding a total runtime of approximately 100 minutes per subpopulation. Simulations and data applications were parallelized across 50 cores to mitigate computational cost. The reported runtimes reflect the final experimental configurations; preliminary experiments required additional compute not included in the totals above.

\end{document}